\documentclass[conference]{IEEEtran}
\IEEEoverridecommandlockouts
\usepackage{cite}

\usepackage{graphicx}
\usepackage{graphicx}
\usepackage{booktabs}
\usepackage{multirow}
\usepackage{array}
\usepackage{makecell}

\usepackage{amsfonts}
\usepackage{amsmath}
\usepackage{amssymb}
\usepackage{wrapfig}
\usepackage{url}
\usepackage{tikz}

\usepackage{algorithm}
\usepackage{algorithmic}
\usepackage{capt-of}

\newcommand\red[1]{{\color{red}#1}}

\usepackage{color, colortbl}

\usepackage{todonotes}
\usepackage{enumitem}

\begin{document}

\title{Evolutionary latent space search for driving human portrait generation}

\author{\IEEEauthorblockN{Benjamín Machín}
\IEEEauthorblockA{\textit{Universidad de la República} \\
Montevideo, Uruguay \\
benjamin.machin@fing.edu.uy}
\and
\IEEEauthorblockN{Sergio Nesmachnow}
\IEEEauthorblockA{\textit{Universidad de la República} \\
Montevideo, Uruguay \\
sergion@fing.edu.uy}
\and
\IEEEauthorblockN{Jamal Toutouh}
\IEEEauthorblockA{\textit{ITIS Software, Universidad de Málaga} \\
Malaga, Spain \\
jamal@lcc.uma.es}
}

\maketitle

\begin{abstract}
This article presents an evolutionary approach for synthetic human portraits generation based on the latent space exploration of a generative adversarial network. The idea is to produce different human face images very similar to a given target portrait. The approach applies StyleGAN2 for portrait generation and FaceNet for face similarity evaluation. The evolutionary search is based on exploring the real-coded latent space of StyleGAN2. The main results over both synthetic and real images indicate that the proposed approach generates accurate and diverse solutions, which represent realistic human portraits. The proposed research can contribute to improving the security of face recognition systems.
\end{abstract}


\onecolumn

\noindent {\Huge IEEE Copyright Notice} \\ \\
{\Large \copyright 2021 IEEE. Personal use of this material is permitted.  Permission from IEEE must be obtained for all other uses, in any current or future media, including reprinting/republishing this material for advertising or promotional purposes, creating new collective works, for resale or redistribution to servers or lists, or reuse of any copyrighted component of this work in other works.} \\

{\Large This article was accepted and presented in the 2021 {IEEE Latin American Conference on Computational Intelligence (LA-CCI)}. Please, cite the paper as: Benjamín Machín, Sergio Nesmachnow, and Jamal Toutouh. Evolutionary latent space search for driving human portrait generation. In \textit{2021 IEEE Latin American Conference on Computational Intelligence}, pages 1--6, 2021.}

\twocolumn

\vspace{-0.9cm}
\section{Introduction}
\vspace{-0.01cm}
%Contexto
Generative Adversarial Networks (GANs) are machine learning methods to learn generative models~\cite{goodfellow2014generative}. Generative models take a training dataset drawn from a specific distribution and learn to represent an estimate of that distribution. 

% Generally, 
GANs consist of two artificial neural networks (ANNs): a generative model (\textit{generator}) and a discriminative model (\textit{discriminator}), which apply adversarial learning to optimize their parameters. 
The generator $g$ learns how to transform input vectors from a random latent space $z$ into “fake” samples $x'$, i.e., $g(z)=x'$, that approximate the true data distribution. Simultaneously, the discriminator learns to distinguish the “real” samples from the training dataset
$x$, from the ones produced by the generator $x'$. GAN training converges to an optimal generator that approximates the real distribution so well that it deceives the discriminator, which randomly labels real and fake samples. 

GANs have demonstrated being a successful tool for many applications that require the creation of synthesized data, especially those concerning multimedia data (e.g., images, sound, and video), healthcare, and other areas~\cite{Pan2019,Toutouh2021, toutouh2020conditional, toutouh2020generative}.

\if 0
%Contexto
Generative Adversarial Networks (GANs) are 
% a set of 
machine learning methods to learn generative models~\cite{goodfellow2014generative}. Generative models take a training dataset drawn from a specific distribution and learn to represent an estimate of that distribution. 

% Generally, 
GANs consist of two artificial neural networks (ANNs): a generative model (\textit{generator}) and a discriminative model (\textit{discriminator}), which apply adversarial learning to optimize their parameters
%. When applying unsupervised GAN training, t
via unsupervised training.
The generator 
$g$
learns how to transform input vectors from a random latent space 
%(prior $z$) 
$z$ 
into 
%“fake/synthesized” 
“fake” 
samples $x'$, 
i.e., $g(z)=x'$,
that approximate the true distribution. Simultaneously, the discriminator learns to distinguish the “real” samples from the training dataset
%$x$, 
 from the ones produced by the generator. GAN training converges to an optimal generator that approximates the real distribution so well that it deceives the discriminator, which 
%only provides a random label for 
randomly labels 
real and fake samples.

\fi

% Motivacion
% Unsupervised GAN training produces generators $g$ that randomly produces fake samples $x'$ from the entire training data distribution given a random vector from the latent space $z$, i.e., $g(z)=x'$. 

In general, the fake sample produced by the generator is determined by the latent space vector read. 
This latent space to produce the samples in a GAN is generally defined by a high dimensional random distribution, e.g., Gaussian distribution, linked to the trained generative model. 

This article presents a proposal for finding a sub-latent space (i.e., vectors from the latent space) that produces fake samples that meet given criteria. 
Specifically, on the basis of a GAN pre-trained to produce realistic human face images, the proposed research aims at finding latent space vectors that make that GAN to produce human portraits with specific attributes. 
In this case, the target human face images 
% that we want 
to create should be similar to a given target human portrait to the human eye.   
The large search space defined by the latent space makes impractical the use of traditional optimization methods for this purpose (e.g., enumeration techniques, backtracking or dynamic programming). Thus, heuristic and metaheuristics~\cite{Nesmachnow2014} are useful methods to perform the search using bounded computational resources.

Thus, in the proposed research, an evolutionary algorithm (EA) is applied to search the latent space for vectors. 
StyleGAN, a well-known GAN pre-trained to randomly create realistic human face images~\cite{Karras2020}, is the one used to generate the tentative samples (human face images). 
FaceNet~\cite{Schroff2015}, an ANN pre-trained to extract the main features of human face images, is employed to evaluate the generated samples and guide the EA during the evolutionary process.
Therefore, the FaceNet output is applied to compute the distance between the target human face image and the generated one. 

FaceNet is widely used by face recognition systems~\cite{william2019face}.
One of the main aims of this research is to synthetically create different human face images that according to FaceNet belong to the same person.
Thus, the proposed approach is useful to create adversarial samples able to deceive face recognition systems based on ANN. This research line can contribute to improve the security of face recognition systems.

This work addresses the capability of generative models to produce samples from the whole data distribution (i.e., create any human face image) and the usefulness of EAs to perform an efficient search of the latent space.
The main contributions of this article are: 
i) a method based on EAs to generate synthesized human portraits to be similar to a given target face, which could be used to deceive face recognition systems, 
ii) the analysis the efficacy of the proposed approach when dealing with different types of target human faces, and {iii)} the analysis of its computational cost.

%Organizacion
The article is organized as follows. Next section introduces the problem of exploring the latent space for portrait generation. Section~\ref{Sec:Meth} describes the methodology and Section~\ref{Sec:EA} presents the proposed EA 
% to solve 
for
the problem. The experimental analysis is reported in Section~\ref{Sec:Exp}.
Finally, Section~\ref{Sec:Conc} presents the conclusions and % formulates 
the main lines for future work.

\section{Exploring latent space for face generation}
\label{Sec:Problem}

This section describes the problem of exploring the latent space of GANs for generating faces and reviews relevant related works.

\subsection{Latent space exploration}

Most of the research on GANs aims at improving their reliability and the accuracy of the trained generative models (i.e., generators). 
Lately, few studies have been focused on the latent space, which is unique for each generator and eventually determines its produced samples. 
Generally, the generator produces an output (e.g., an image) by randomly sampling the latent space distribution, i.e., by taking a random vector from the latent space. 
The latent space is defined by a high dimensional random distribution called the prior. 
This article focuses on the latent space exploration problem, which consists in finding the sub-spaces of the latent space (or computing vectors from the prior) for forcing a given generator to produce a specific output (i.e., samples that meet certain criteria). 

Specifically, the proposed research addresses the latent space exploration of a given pre-trained generator (StyleGAN) for creating human face images similar to a target portrait. 
Due to the high dimensionality of the latent space, the search es carried out by using an EA.
The sample produced by StyleGAN is evaluated by FaceNet, another pre-trained ANN, which evaluates the similarity between the produced sample and the target portrait. 
The output computed by FaceNet is used to guide the EA search through the latent space.

\subsection{Related work}
\label{sec:related-work}

%In this research, evolutionary computation (EC) is applied to evolve latent vectors to generate synthetic images. <= NO SE ENTIENDE, no se puede empezar una subsección con "This", introduce un término nuevo (EC) que no se usa en todo el artículo, habla de "synthetic images" cuando quedamos que el artículo es sobre faces.

% LARGO,q ueda para revista
% One of the first LVE approaches to create images was focused on the generation of fingerprints (\textit{Synthetic MasterPrints})~\cite{bontrager2017deepmasterprint}. The authors compared four metaheuristics (Hill-Climbing, Covariance Matrix Adaptation Evolution Strategy, Differential Evolution, and Particle Swarm Optimization) to search latent space vectors to optimize a metric proposed by them, the Modified Marginal Success Rate. This study presented a two-stage methodology similar to the one proposed in this study, i.e., the use of unsupervised training of GANs and second the evolution of latent space. The main difference to our approach is that they train their model and, in this research, a pre-trained model is used. 
% CORTO
The Latent Variable Evolution (LVE) approach was originally proposed for the generation of fingerprints~\cite{Bontrager2018}, using metaheuristics to optimize an ad-hoc quality metric.
%
%Some research has already been made in the exploration of the latent space that not apply evolutionary approaches. For example, Generative Kernel Principal Component Analysis models were proposed to explore the latent space allowing to move along component in the feature space allows for the interpretation of components and consequently additional insight into the underlying latent space~\cite{winant2020}.
Non-evolutionary approaches have also been applied to explore the latent space of GANs. For example, models based on Generative Kernel Principal Component Analysis  allow for the interpretation of components by moving 
% along the components 
in the feature space, thus providing an 
% additional 
insight into the underlying latent space~\cite{winant2020}. 
% \red{intenté mejorar la redacción de esto que tenía algún errorcito pero no sé si está corecto}

%which allows moving along components in the feature space allows for the interpretation of components and consequently additional insight into the underlying latent space~\cite{winant2020}.

%Focusing on human face images, EAs have been applied to search latent space to improve the diversity of the generated samples~\cite{Fernandes2020}. In this research??????, the authors trained an unsupervised GAN by using Facity dataset. Then, a genetic algorithms or a Map Elites method evolves solutions that represents a set of 50~latent space vectors to generate 50~images. The fitness function is evaluated in terms of the diversity of the produced faces. The research proposed here??? each solution represents a unique latent space vector used to create one sample of a pre-trained StyleGAN2 model.  

Focusing on human face images, Fernández et al.~\cite{Fernandes2020} applied EAs to search the latent space to improve the diversity of samples generated by an unsupervised GAN.
%using the Facity dataset. 
A genetic algorithm and a Map Elites method were used to evolve solutions that represent a set of 50 latent space vectors to generate 50 images. The fitness function was evaluated in terms of the diversity of the produced faces. 

StyleGAN was proposed as an improvement over deep convolutional GANs, 
%(DC-GANs) in which
a model where
the generator embeds the input latent vector into an intermediate latent space, which has an important effect on how the variation factors are represented in the ANN~\cite{Karras2020}. 
%StyleGAN automatically learns, unsupervised separation of high-level attributes (e.g., pose and identity) and stochastic variation in the generated images (e.g., hair). 
StyleGAN automatically learns the separation of high-level attributes (e.g., pose and identity) and the stochastic variation in the generated images (e.g., hair). 
It allows smooth interpolation and style mixing with high quality output images. 

Shen et al.~\cite{Shen2020} proposed InterFaceGAN for semantic face editing by interpreting the latent semantics learned by GANs, and studied how different synthetic face semantics are encoded in the latent space. 
InterFaceGAN finds hyperplanes that divide the latent space in subregions that generate specific attributes. By leveraging those regions, 
% the authors were able to manipulate in an isolated manner, facial attributes for gender, age, presence of eyeglasses, smile and pose on the images generated by PGGAN and StyleGAN. 
InterFaceGAN manipulates in an isolated manner facial attributes for gender, age, presence of eyeglasses, smile, and pose on images generated by 
% PGGAN and >- NO ESTÁ DEFINIDO
StyleGAN. 
%\red{They\textbf{??}} also applied these transformations to real images by projecting original images from real people into the latent space and then applying \red{their\textbf{??}} method.
%InterFaceGAN was proposed to allow semantic face editing by interpreting the latent semantics learned by GANs~\cite{Shen2020}. The authors studied how different synthetic face semantics are encoded in the latent space. InterFaceGAN finds hyperplanes that divide regions of the latent space in subregions that generate a specific attributes is presented. By leveraging those regions, they were able to manipulate in an isolated manner, facial attributes for gender, age, presence of eyeglasses, smile and pose on the images generated by PGGAN and StyleGAN. They also applied these transformations to real images by projecting original images from real people into the latent space and then applying their method.

%StyleGAN2 Distillation architecture, the model used in our approach, was proposed to change the facial features of gender and age, and to perform style transfer and image morphing~\cite{Viazovetskyi2020}. The distillation allows to extract the information about faces’ appearance and the ways they can change (e.g. aging, gender swap) from StyleGAN into image-to-image network. 
The StyleGAN2 Distillation architecture was proposed to change the facial features of gender and age, and to perform style transfer and image morphing~\cite{Viazovetskyi2020}. 
The distillation allows extracting the appearance information from the generated faces and provides a way to manipulate these attributes.

All reviewed works trained their own architecture and focused on the exploration of the latent space to modify specific facial attributes (gender, age, pose, etc.). In contrast, the approach proposed in this article does not require to train any new ANN model and leverages on the use of pre-trained models for both, generation and evaluation of the images. 
Additionally, the main goal is to generate faces that are similar to a given one, being able to deceive the model used to evaluate the similarity between the samples. 

%Besides, the main goal is to generate faces as similar to a given one which is able to deceive the model used to evaluate the similarity between the samples. 

\section{Methodology for automatic generation of human faces}
\label{Sec:Meth}

%Method \red{incluir título descriptivo}}
%This section describes the research methodology applied for portrait generation. 
%used and how it was instantiated using pre-existing models.   

%\subsection{Overview}
%\red{Our approach is simple} and consists of evolving a population of latents using a metric provided by another model as measure of \red{fitness \textbf{no podemos hablar de fitness en esta sección, recién se presenta en la sección siguiente. CORREGIR}}. Theoretically, any model that takes an image as input and outputs a score can be used to this end. We test the idea with a face generation GAN and a pre-trained facial recognition model, with the goal of finding latents that when passed through the generator, will output a face that resembles a face from an existing person, according to the metric provided by the facial recognition model.

%This section describes the methodology applied for the automatic generation of human face images similar to a given target face image by using EAs and two pre-trained ANN models: a GAN to generate the images and a facial recognition model to assess the images generated by the GAN (see Fig.~\ref{Fig:methodology}). 
%
%This section describes the methodology applied for the automatic generation of human face images similar to a given target face image by using EAs and two pre-trained ANN models: a GAN to generate the images and a facial recognition model to assess the images generated by the GAN (see Fig.~\ref{Fig:methodology}). 
A specific methodology is proposed for the automatic generation of human face images, with the main goals of generating high-quality, similar to the target face, but also with diversity while maintaining the main features of the target face. An hybrid methodology is applied, combining an EA and two pre-trained ANN models: a GAN to generate the images and a facial recognition model to assess the quality and diversity of generated images. 

A population of latent vectors is evolved, which are given as input to a generative model to synthesize human face images. 
%The search is guided according to a quality indicator provided by another ANN model. The quality of the produced faces is assessed according to their similarity with the target image.
The search is guided by a quality indicator provided by another ANN model, that indicates the similarity of generated faces to the target image.

% The generative model used is StyleGAN2 \cite{Karras2019}, 
% which was the state of the art in face generation at the time of this work. 
% a state-of-the-art GAN for face generation.
StyleGAN2 \cite{Karras2020}, a state-of-the-art GAN for face generation, is used as generative model.
% a state-of-the-art GAN for face generation.
The model was pre-trained on the Flickr-Faces-HQ Dataset 
%(FFHQ) dataset provided by NVIDIA at \hbox{\url{https://github.com/NVlabs/ffhq-dataset}}. 
by NVIDIA.
% (\hbox{\url{https://github.com/NVlabs/ffhq-dataset}}). 
StyleGAN2 provides higher resolution image generation and the disentanglement of the latent space, which allows style mixing and smooth interpolation. 

In turn, the quality of images is assesed using FaceNet~\cite{Schroff2015}, a well-known face recognition model that transforms 
%an aligned 
a
face image into data points in a high dimensional space, i.e., 
%FaceNet returns 
a vector of 128 continuous values for each input image. The output vectors of different input images, which belong to the same face, have the property of being geometrically close to each other. Thus, FaceNet is used to assess the resemblance of the generated faces to the target face.

\begin{figure*}[!h]
\setlength{\belowcaptionskip}{-20pt}
\setlength{\abovecaptionskip}{3pt}
    \centering
    \includegraphics[trim=0.15cm 12.2cm 4.5cm 0, clip, width=\textwidth]{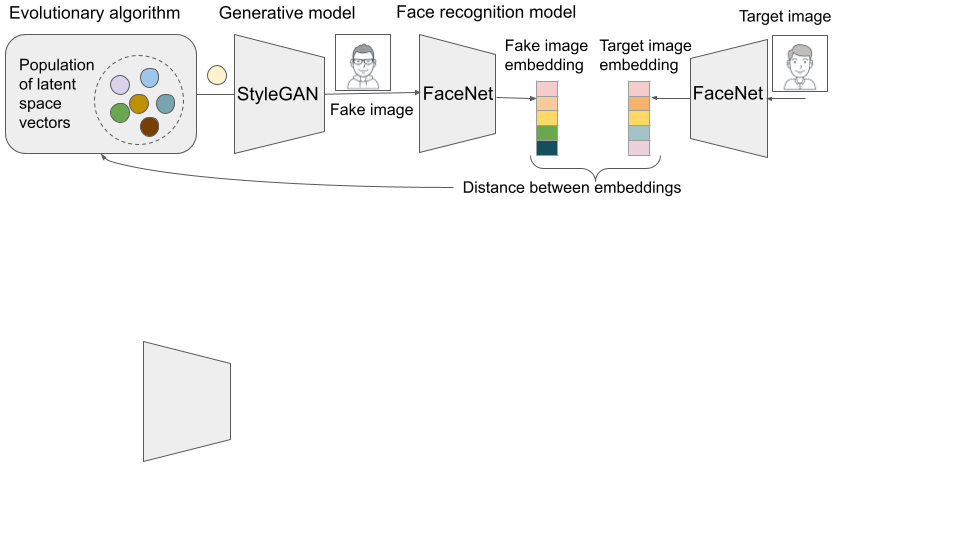}
    \caption{Methodology proposed to generate human face images close to a target one.}
    \label{Fig:methodology}
\end{figure*}

Fig.~\ref{Fig:methodology} describes the modules integrated in the proposed system and their interactions.

\section{An evolutionary algorithm for latent space exploration}
\label{Sec:EA}

%A genetic algorithm is used to perform the latent space exploration in a guided manner, as described in the overview. In this subsection we describe the main components of our approach. 
This section describes the proposed EA for exploring the latent space of GANs applied to realistic portrait generation.

%\subsubsection{Individuals representation}
\subsubsection{Solution encoding}
%The representation of individuals is straightforward in our case, since latent vectors in the StyleGAN2 space are simply vectors of 512 floating point numbers, and the space has properties that facilitate the crossing and mutation operations described in the following sections.
The applied solution encoding considers a vector of 512 floating point numbers, representing the components of the latent vectors in the StyleGAN2 space~\cite{Karras2020}. Values in the latent space during the training of StyleGAN2 were sampled from a standard normal distribution. %\red{VER: en el informe dice centrados en cero y escalados} 

\subsubsection{Fitness function}
%In each step of the evolutionary process, we calculate the fitness value for each latent vector in the population. First, we get the corresponding image using the StyleGAN's generator, and compare it to a target face, for which we have previously obtained an embedding representation. To get embeddings from the generated image, we perform face detection and alignment and then we extract the embeddings using the FaceNet model. The fitness of each individual is the distance of its embeddings to the embeddings of the target image. \red{voy a agregar un diagrama para esto}
The fitness function evaluates the similarity of the target face to the face generated by StyleGAN2 using the individual as input. 
Embeddings, which represent the main features of a human face image, are used characterize the face generated. 
These embeddings are obtained by a forward pass through the facial recognition model, as described in Fig.~\ref{Fig:methodology}.
The fitness value of an individual is the opposite value of the Euclidean distance (L2 norm) of the generated face embeddings to the target image embeddings. 
%This is illustrated in Fig.~\ref{Fig:fitness}.
%
An initial implementation considering the evaluation of one individual at a time resulted in a very inefficient search procedure, mainly due to the significant inference times of the 
%neural networks used, 
used ANNs
% , which were ran once per individual, missing to use the networks' batch processing capabilities. 
when batch processing capabilities are not used.
% \red{completar} \blue{listo - ajusten redacción si lo creen necesario}. 
In order to improve the computational efficiency of the proposed EA, a parallel model for evaluation using batches was implemented. 

\subsubsection{Evolutionary operators} The proposed evolutionary operators are described next.

\paragraph{Initialization} A random initialization operator was applied. Each value in a solution encoding is generated according to a normal distribution $\mathcal{N}(0,1)$.
%\red{Benjamín, completar, los detalles no están en el informe}\blue{al final usé una normal estándar nomás, hay que poner algo más? sin dudas esto hay que explorarlo más también} \red{hay que poner media y sigma, lo pongo}. 
Seeding the population is a valuable idea in order to guide the search towards specific regions of the space, thus promoting certain features in the generated faces. 
%Esto no iria aquí, puesto que se propone es una búsqueda dentro de la búsqueda. Y ademas, es un trabajo futuro. 
This alternative is proposed as one of the main lines for future work.
% in order to generate ad-hoc faces.

\paragraph{Selection} The tournament selection was applied, which provides a good selection pressure to guarantee diversity during the search. Three individuals participate in the tournament, and one survives. This parameterization was determined in preliminary calibration experiments.

\begin{figure}[!h]
    \centering
    \includegraphics[width=0.9\linewidth]{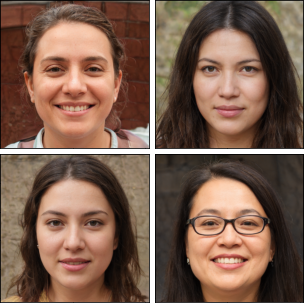}
    \caption{Two examples of the phenotypes resulting of the application of the BLX-$\alpha$ recombination operator
    %. The upper images are the original images, below the resulting images of the recombination.
    (top: parents, bottom: offspring).
    }
    \label{Fig:sample_recombination}
\end{figure}

%\subsubsection{Crossing}
\paragraph{Recombination}
%\red{Different strategies were tested to perform crossing of individuals, which simply reduces to the decision of how to create two new latents from the two parent ones...}
%
The blend-alpha crossover operator (BLX-$\alpha$)~\cite{Eshelman1993} was applied, mainly because its search pattern adapts to the latent space exploration. The main idea is to properly exploit the intervals determined by values encoded in both parents and not focusing on simple combinations of encoded values. BLX-$\alpha$ uniformly selects values between two points that contain the two parents, but may extend equally on either side determined by the $\alpha$ parameter. This is a common procedure in real-encoding EAs to solve different problems.
In preliminary calibration experiments, BLX-$\alpha$ allowed the proposed EA to compute better results than standard crossover operators. The value of $\alpha$ must be set to guarantee a proper exploitation of solutions. This procedure is often performed empirically.
Fig.~\ref{Fig:sample_recombination} presents examples of the phenotypes resulting of the application of the proposed recombination operator.

\begin{figure}[!h]
    \centering
    \includegraphics[width=0.9\linewidth]{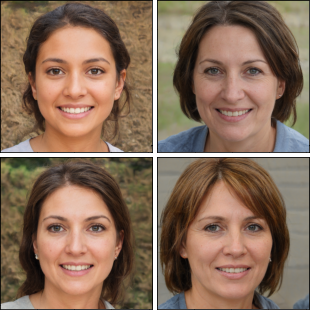}
    \caption{Two examples of the phenotypes resulting of the application of the proposed mutation operator 
    %(on the top, the original images; and below them the results after mutation).
    (top: original image, bottom: mutated image).
    %A mutation probability of 0.1 was considered for each element in the encoding.}
    }
    \label{Fig:sample_mutation}
\end{figure}

\paragraph{Mutation} Several mutation operators were studied in preliminary experiments to determine a proper means to introduce diversity in the search. The best results were obtained applying a Gaussian mutation with mean $\mu = 0$ and standard deviation $\sigma = 
%$\red{VER, en el informe dice 1, es correcto? (parece muy amplia la dispersión) \blue{si, puse 1 en las pruebas porque no parecía variar mucho la cosa. este es uno de los puntos que hay que explorar más}}
1$, which empirically showed to generate appropriate diversity without being too disruptive. Fig.~\ref{Fig:sample_mutation} presents some examples of the phenotypes resulting of the application of the proposed mutation operator.

\subsubsection{Implementation details}
The proposed EA was implemented in Python 3.7 using the DEAP library version 1.3.1. 
Several improvements were implemented to accelerate the fitness evaluation process, including:
\begin{enumerate}[label=\Alph*]
    \item \textit{Batch fitness evaluation}. By default, the fitness evaluation in DEAP is performed sequentially, i.e., one individual at a time, mapping the fitness function to the list that holds the population. 
    %On the other hand, performing calculations in batches is a common technique when dealing with complex evaluation functions in EAs and when working with large ANNs. In our case, both the face generation and the face recognition models can be executed in batches. In order to be able to exploit these batching capabilities, the source code of DEAP's evolutionary algorithm was modified to evaluate the entire population at once, leaving the partitioning in batches up to the particular settings of the individual models. This simple modification accelerated each evolutionary step twenty-fold.
    This procedure demands a large computation time when working with large ANNs.
    In order to speed up the fitness evaluation, both the face generation and the face recognition models can be executed in batches. This is a common technique when dealing with complex evaluation functions in EAs~\cite{Alba2012}. The source code of DEAP was modified to exploit batching capabilities, for evaluating the entire population at once.
    The batch partitioning is left to the particular settings of each model. 
    %\red{This simple modification accelerated each evolutionary step twenty-fold \textbf{esto es incomprensible, propongo sacarlo}}.
    %\item \textit{Image size reduction}. Since the facial recognition model is trained on images of 160 pixels width and height, the generated images, of 1024 pixels width and height, were downscaled before detection, alignment and embeddings generation, yielding a significant acceleration of the fitness evaluation process. Note that this is applied only in the fitness calculations, the final output of preserves its full resolution.
    \item \textit{Image size reduction}. The facial recognition model was trained on images of size 160$\times$160 pixels. 
    % (width and height). 
    In consequence, when computing the fitness function, all generated images 
    (1024$\times$1024 pixels) were downscaled before detection, alignment, and embeddings generation, yielding a significant acceleration of the fitness evaluation process. The size reduction was applied only in the fitness calculations. Thus, the final output preserves its full resolution.
    %\item \textit{Replacement of the face detection algorithm}. During the implementation, it was noticed that the coordinates of the bounding boxes returned by the face detection model were more or less constant. We understand this is a property of the faces generated by StyleGAN2. To test this hypothesis, a set of 10,000 random images were generated and their mean coordinates were analyzed. The interquartile range for the coordinates was of 8 pixels in average, which should not affect the facial recognition model's performance.    
%    \item \red{\textit{Replacement of the face detection algorithm}. During the implementation, it was noticed that the coordinates of the bounding boxes returned by the face detection model were more or less constant. We understand this is a property of the faces generated by StyleGAN2. To test this hypothesis, a set of 10,000 random images were generated and their mean coordinates were analyzed. The interquartile range for the coordinates was of 8 pixels in average. After this validation, the detection model was replaced by a fixed set of coordinates, further reducing the fitness calculation time. This change should not affect the facial recognition model's performance.}
    \item \textit{Replacement of the face detection algorithm}. 
    The facial recognition model requires 
    % faces to be cropped tightly 
    a tight crop
    to work properly. For this purpose, a Multitask Cascaded Convolutional Network (MTCNN)~\cite{Zhang2016} was used. % in our first implementation. 
    A preliminary validation analysis detected that, when applied to images generated by StyleGAN2, the coordinates of the bounding boxes returned by MTCNN do not vary significantly. This hypothesis was confirmed 
    % in preliminary experiments by generating a set of 10,000 random images and analyzing the results. 
    in experiments that analyzed 10,000 generated images: results demonstrated that the average interquartile range of the bounding box coordinates was 8 pixels (below 4\% of deviation). Thus, instead of using MTCNN to crop the faces, a fixed average bounding box was used, removing the detection step from the fitness calculations, and
    further reducing the execution times.
    % should not affect the facial recognition model's performance.}
\end{enumerate}

\section{Experimental analysis}
\label{Sec:Exp}

This section reports the experimental evaluation of the proposed evolutionary search of the latent space in terms of the quality and diversity of the generated images, and the computational efficiency of the search. This experiments were performed on the Colaboratory platform from Google (\url{colab.research.google.com/}). 

\subsection{Problem instances}

\begin{figure}[!h]
\centering
     \includegraphics[width=\linewidth]{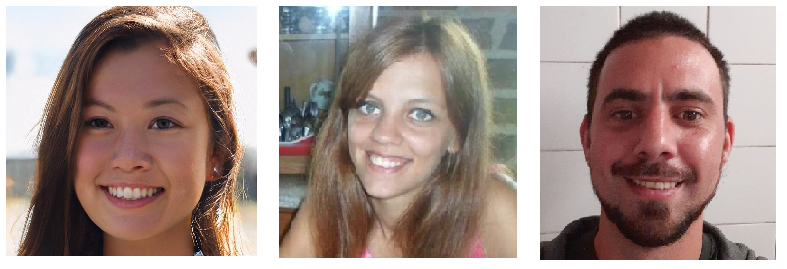}
     \includegraphics[width=\linewidth]{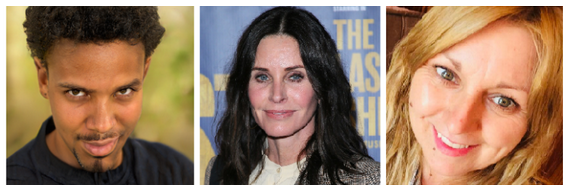}
    %\caption{Instances used in configuration (top) and validation (bottom) experiments. Top (from left to right): G1, W1 and M1. Bottom (from left to right): G2, W2 and W3.}
    \caption{Instances used in configuration experiments (top: G1, W1 and M1) and validation experiments (bottom: G2, W2 and W3).}
    \label{Fig:valid_instances}
\end{figure}

Six problem instances were taken into account by using the six different target human portraits shown in Fig.~\ref{Fig:valid_instances}:  
G1 and G2 are images randomly produced by StyleGAN; 
W1, W2, and W3 are real images of women; and M1 is a real image of a man. %
The use of these different types of images allows the evaluation of the proposed approach when dealing with synthesized/real images and women/men faces.

G1, W1, and M1 were used for parameter setting experiments and G2, W2, and W3 validation experiments.

\subsection{Parameters setting experiments}

Parameters setting experiments were performed on problem instances G1, M1, and W1 to find the most appropriate configuration of the recombination ($p_R$) and mutation ($p_M$) probabilities. The population size (200 individuals) and the number of generations (500) were established in previous experiments.    
Considered candidate values were $p_R \in \{0.6, 0.75, 0.9\}$ and $p_M \in \{0.001, 0.01, 0.1\}$. The nine combinations of candidate values were studied, performing 30 independent executions of each parameter configuration for the considered instances. 

\subsection{Validation results}

This subsection reports and discusses the results of the 30 independent executions of the proposed EA configured according to the results of the parameters settings experiments (i.e., population size = 200, number of generations = 500, $p_R$~=~0.75, $p_M$ = 0.001, $\alpha$ = 0.2). 
Four aspects are 
%taken into account in the study: 
analyzed in validation experiments: 
the 
%fitness/distance 
fitness
results and its evolution, the computational cost of the proposed EA, and the quality and diversity of the generated face images. 

\textit{Fitness values.} 
Table~\ref{Tab:validation} reports the minimum, mean, and standard deviation of the computed distances (opposite fitness values, i.e., lower values are better). 
For the three instances, the EA converged to lower values than the ones computed during the parameter setting experiments. 
When comparing the results of the three instances, 
the proposed approach provides better results when dealing with synthesized target images, i.e., G2, than when dealing with real ones. 
Fig.~\ref{Fig:generations} illustrates the evolution of the minimum distance between the generated images and the target face (opposite fitness values) of ten independent runs for instance W2.

% The parameters and configurations found during the previous experiments were validated against the set of instances shown in Fig.~\ref{Fig:valid_instances}. Again, 30 independent executions were performed for each instance. The numerical results are shown in table \ref{Tab:validation}. The bigger execution times for W3 are directly related to the hardware allocation mechanism of Google Colab, which randomly chooses between different types of hardware instances. Each instance's validation was executed in a single session (with fixed hardware for the instance's validation executions).

% \begin{table}
% \centering
% \caption{Validation results (fitness and execution time) on 30 independent runs.}\label{Tab:validation}
% \begin{tabular}{|c|c|c|c|}
% \hline
% Instance & Min. Distance & Distance & Exec. Time\\
% \hline
% G2 & $0.350$ & $0.453$\,$\pm$\,$0.041$ & $625$\,$\pm$\,$8s$\\
% \hline
% W2 & $0.550$ & $0.655$\,$\pm$\,$0.049$ & $642$\,$\pm$\,$6s$\\
% \hline
% W3 & $0.420$ & $0.495$\,$\pm$\,$0.041$ & $905$\,$\pm$\,$6s$\\
% \hline
% \end{tabular}
% \end{table}

\begin{table}[!h]
\vspace{-0.1cm}
\setlength{\abovecaptionskip}{0pt}
\setlength{\belowcaptionskip}{0pt}
\renewcommand{\arraystretch}{0.95}
\centering
\setlength{\tabcolsep}{6pt}
\caption{Validation results (distance values and execution times).}
\label{Tab:validation}
\begin{tabular}{l r r r r rrr}
\toprule 
\multicolumn{1}{c}{\multirow{2}{*}{\textit{instance}}} & \phantom{ab}  &
\multicolumn{2}{c}{\textit{distance}} &\phantom{ab}  &
\multicolumn{2}{c}{\textit{execution time (s)}} \\
\Xcline{3-4}{1\arrayrulewidth}  \Xcline{6-7}{1\arrayrulewidth}  
&  & min & mean\,$\pm$\,std && min & mean\,$\pm$\,std \\
\midrule
G2 && $0.350$ & $0.453$\,$\pm$\,$0.041$ && 614 & $625$\,$\pm$\,$8$\\
W2 && $0.550$ & $0.655$\,$\pm$\,$0.049$ && 630 & $642$\,$\pm$\,$6$\\
W3 && $0.420$ & $0.495$\,$\pm$\,$0.041$ && 896 &$905$\,$\pm$\,$6$\\
\bottomrule
\end{tabular}
\end{table}

\begin{figure}[!h]
\vspace{-0.5cm}
\setlength{\belowcaptionskip}{-12pt}
\setlength{\abovecaptionskip}{-3pt}
    \centering
     \includegraphics[width=0.95\linewidth]{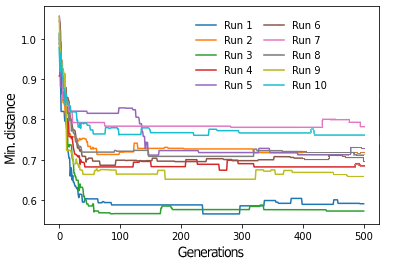}
    \caption{Evolution of distance values for ten independent runs on W2 instance.}
    \label{Fig:generations}
\end{figure}

\textit{Computational cost.} 
Table~\ref{Tab:validation} also reports the execution times (minimum, mean, and standard deviation) of the proposed EA. The execution times of the optimized implementation were approximately seven times faster than the initial implementation without the improvements 
%mentioned in Section~\ref{Sec:EA}. 
to accelerate the fitness function calculation.
% As it can be seen, the proposed approach is able to provide images close to the target one in less than 15~minutes. This is one of the main advantages of using pre-trained models, there is no need of training complex ANN models (that imply high computational costs) to get competitive results.   
The proposed approach was able to compute accurate images (close to the target) in less than 15 minutes, showing the main advantage of using a pre-trained model: there is no need of training complex ANN models (that imply high computational costs) to get competitive results.   
The larger execution times for W3 are directly related to the hardware allocation mechanism of Google Colab, which randomly chooses between different types of hardware instances. 
% Each instance's validation was executed in a single session (with fixed hardware for the instance's validation executions). <- ESTO NO SE ENTIENDE

\begin{figure}[!h]
\centering
     \includegraphics[width=\linewidth]{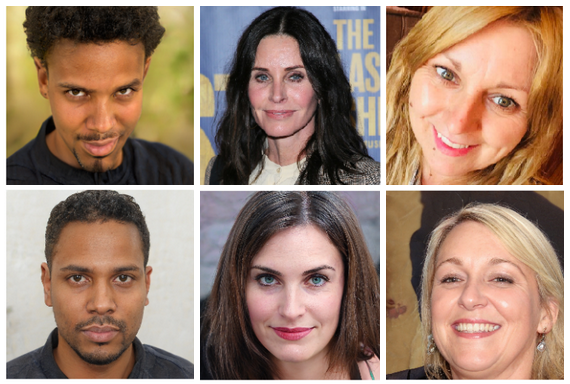}
    \caption{Validation instances (top: G2, W2 and W3) and their associated best solution found in the validation experiments (bottom: SG2, SW2 and SW3).}
    \label{Fig:validation_res}
\end{figure}

\textit{Solution quality.} 
The quality of results was evaluated according to the capability of the proposed method to deceive FaceNet. The rationale behind the proposed evaluation methodology is to compare the embeddings provided by FaceNet 
%for the image produced by the best solution found 
to the best generated image shown in Fig~\ref{Fig:validation_res}. 
and 
%\red{for a different image of the same target person with the embeddings of the target person. \textbf{no está bien explicado}} 
to a different image of the same target person.
%Table~\ref{Tab:face_recognition} shows this comparison for the instances W2 and W3 by presenting the $L_2$ distance between the embeddings of the target image and the best solution (second column) and the $L_2$ distance between the embeddings of the target image and another image of the same target person (third column). 

Table~\ref{Tab:face_recognition} reports the comparison of the computed embeddings with FaceNet for instances G2, W2, and W3, considering the $L_2$ distance as relevant indicator. As stated, the baseline result is the $L_2$ distance between the embeddings of the target image and another image of the same target person. For both instances, the distance between the synthesized image and the target one is shorter than the baseline value (up to 27.9\% of reduction was achieved for problem instance W3). 
%These results show that our method fooled the facial recognition model, being able to generate synthetic faces that produce embeddings closer to the target person than embeddings obtained from other image of the same person.

The reported results show that the proposed method was able to deceive the facial recognition model, generating synthetic faces that produce embeddings closer to those of the target person {than the embeddings obtained from a different image of the same person}.

\begin{table}[!h]
\setlength{\tabcolsep}{6pt}
\renewcommand{\arraystretch}{1.0}
\centering
\caption{Quality results in terms of distance between embeddings} % for W2 and W3.}
\label{Tab:face_recognition}
\begin{tabular}{l r r r}
\toprule 
\textit{instance} & \textit{target vs.~fake} & 
% \textit{target vs.~other same real} \\
\textit{baseline} & \multicolumn{1}{c}{$\Delta$} \\
\midrule
G2 & $0.350$ & - & - \\ 
W2 & $0.550$ & $0.679$ & 12.9\%\\
W3 & $0.420$ & $0.583$ & 27.9\%\\
\bottomrule
\end{tabular}
\end{table}

\begin{figure}[!h]
    \centering
    \includegraphics[width=\linewidth]{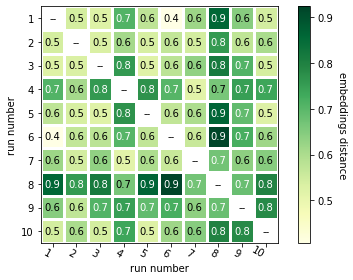}
    \caption{Pairwise distances for ten solutions for validation instance W2.}
    \label{Fig:validation_heatmap}
\end{figure}

\textit{Diversity.} To evaluate the diversity of the generated images, Table \ref{Tab:diversity} reports the pairwise $L_2$ distances between the embeddings of the ten solutions found for each studied instance. 
%\ref{Tab:diversity} reports he summarized results. 
%Pairwise distances between solutions are greater \red{than the distance from the target image to the minimum shown in Table \ref{Tab:validation}\textbf{?}}.
% The mean values of pairwise distances between solutions are greater than 
% To illustrate this, ????
Results suggest a proper robustness of the proposed method.
In turn, the heatmap in Fig.~\ref{Fig:validation_heatmap} presents a representative example of the distances compute for ten solutions of the validation instance W2. Similar results were computed for the other studied validation instances.

\begin{table}[!h]
\renewcommand{\arraystretch}{1.0}
\centering
\setlength{\tabcolsep}{8pt}
\caption{Diversity results in terms of pairwise $L_2$ distances between embeddings} % of solutions for G2, W2 and W3.}
\label{Tab:diversity}
\begin{tabular}{l r r r r rrr}
\toprule 
instance && min & max & mean\,$\pm$\,std\\
\midrule
G2 && $0.314$ & $0.599$ & $0.491$\,$\pm$\,$0.060$\\
W2 && $0.482$ & $0.865$ & $0.645$\,$\pm$\,$0.099$\\
W3 && $0.401$ & $0.674$ & $0.544$\,$\pm$\,$0.070$\\
\bottomrule 
\end{tabular}
\end{table}

% \red{The diversity is shown by the fact that for most of the pairs the distance between them exceeds the minimum distance to the actual target image (as shown in Table \ref{Tab:validation}). The diversity is shown by the fact that for most of the pairs the distance between them exceeds the minimum distance to the actual target image (as shown in Table \ref{Tab:validation} \textbf{no lo entiendo}}

\newpage

%\red{lo voy a redactar} \blue{It is important to remark the diversity in our results because it shows the capability of the proposed method to explore the high dimensional latent space in a way that it can provide a number of different samples that meets the defined criteria, i.e., it can provide diverse human faces that can deceive the  facial  recognition  model.}
The obtained diversity in the computed results is a remarkable feature of the proposed method. It shows the capability of the evolutionary approach to explore the high dimensional latent space in such a way to provide a number of different samples that meets the defined criteria, i.e., it can provide diverse human faces that can deceive the base  facial recognition  model.

\section{Conclusions}
\label{Sec:Conc}

This article presented an evolutionary approach for synthetic human faces generation based on the GAN latent space exploration. 
A specific methodology was proposed to produce human portraits close to a target one (i.e., with specific attributes), by using a pre-trained StyleGAN genertor. 
A specific EA was implemented to explore the latent space, which used a face recognition model, FaceNet, to guide the search. 
In turn, several improvements were included in order to speed up the search. 

The main experimental results 
indicate that the proposed EA was able to generate accurate face images. The capability of the proposed method to deceive FaceNet, the ANN used for similarity evaluation on most of the face recognition systems, was confirmed by the reduced distance between the synthesized image and the target one, which improved up to 27.9\% the baseline result (a different image of the same target person). Diversity results suggest a proper robustness of the proposed method, which was capable of providing a set of different face images similar to the target one.  

The main lines for future work are related to exploring strategies for guiding the search via population seeding or ad-hoc evolutionary operators to generate images with specific characteristics and analyze the capabilities of the proposed method to deceive automatic detection tools as suggested by the quality results.

\vspace{-3pt}

\section*{Acknowledgment}
This research was partially funded by European Union’s Horizon 2020 research and innovation program under the Marie Skłodowska-Curie grant agreement
No 799078, under H2020-ICT-2019-3, and under TAILOR ICT-48 Network (No 952215). %, and by the Universidad de Málaga, Consejería de Economıa y Conocimiento de la Junta de Andalucía and FEDER under grant number UMA18-FEDERJA-003.

\bibliographystyle{IEEEtran}
\bibliography{bibliography}

\end{document}